\documentclass[letterpaper]{article} 
\usepackage{aaai24}  
\usepackage{times}  
\usepackage{helvet}  
\usepackage{courier}  
\usepackage[hyphens]{url}  
\usepackage{graphicx} 
\usepackage{natbib}  
\usepackage{caption} 
\usepackage[utf8]{inputenc} 
\usepackage{booktabs}       
\usepackage{amsfonts}       
\usepackage{nicefrac}       
\usepackage{microtype}      
\usepackage{xcolor}         
\usepackage{amsmath}
\usepackage{soul}
\usepackage{amsthm}
\usepackage{enumitem}
\usepackage{arydshln}
\usepackage{multirow}
\usepackage{algorithm}
\usepackage{algorithmic}
\usepackage{newfloat}
\usepackage{listings}

\usepackage{tikz}
\usetikzlibrary{decorations.pathreplacing,calc}
\newcommand{\tikzmark}[1]{\tikz[overlay,remember picture] \node (#1) {};}

\DeclareCaptionStyle{ruled}{labelfont=normalfont,labelsep=colon,strut=off} 
\floatstyle{ruled}
\newfloat{listing}{tb}{lst}{}
\floatname{listing}{Listing}
\title{IVP-VAE: Modeling EHR Time Series with Initial Value Problem Solvers}
\author{
    Jingge Xiao\textsuperscript{\rm 1}, 
    Leonie Basso\textsuperscript{\rm 1}, 
    Wolfgang Nejdl\textsuperscript{\rm 1}, 
    Niloy Ganguly\textsuperscript{\rm 2}, 
    Sandipan Sikdar\textsuperscript{\rm 1}
}
\affiliations{
    \textsuperscript{\rm 1}L3S Research Center, Leibniz Universität Hannover\\
    \textsuperscript{\rm 2}Indian Institute of Technology Kharagpur\\
    \{xiao, basso, nejdl, sandipan.sikdar\}@L3S.de, niloy@cse.iitkgp.ac.in
}

\begin{document}

\maketitle

\begin{abstract}
Continuous-time models such as Neural ODEs and Neural Flows have shown promising results in analyzing irregularly sampled time series frequently encountered in electronic health records. Based on these models, time series are typically processed with a hybrid of an initial value problem (IVP) solver and a recurrent neural network within the variational autoencoder architecture. Sequentially solving IVPs makes such models computationally less efficient. In this paper, we propose to model time series purely with continuous processes whose state evolution can be approximated directly by IVPs. This eliminates the need for recurrent computation and enables multiple states to evolve in parallel. We further fuse the encoder and decoder with one IVP solver utilizing its invertibility, which leads to fewer parameters and faster convergence. 
Experiments on three real-world datasets show that the proposed method can systematically outperform its predecessors, achieve state-of-the-art results, and have significant advantages in terms of data efficiency.
\end{abstract}

\section{Introduction}\label{sec:intro}

Electronic Health Record (EHR) data contains multi-variate time series of patient information, such as vital signs and laboratory results, which can be utilized to perform diagnosis or recommend treatment \cite{mcdermott2021comprehensive}. The data in EHR time series is often irregularly sampled (i.e., unequal time intervals between successive measurements) and can have missing values \cite{Zhang2022:raindrop}. The irregularity is caused mainly due to unstructured manual processes, event-driven recordings, device failure, and also different sampling frequencies across multiple variables \cite{weerakody2021review}. These complexities make learning and modeling clinical time series data particularly challenging for classical machine learning models~\cite{Shukla2020:survey, sun2020review_ehr_timeseries}. In recent years, significant progress has been made in the development of models for handling irregularly sampled time series data~\cite{Che2018:gru-d,Rubanova2019:latent-ode,Shukla2021:mTAN,Zhang2022:raindrop}, which have been  extensively tested on EHR data.

Neural ODEs \cite{Chen2018:neural-ode} are continuous-time models based on ordinary differential equations (ODEs) that can naturally handle irregularly sampled data. The data is assumed to be generated by a continuous process that is modeled using ODEs. 

\citet{Rubanova2019:latent-ode} further extend the idea and develop Latent-ODE by integrating Neural ODEs and recurrent neural network (RNN) into a variational autoencoder (VAE)~\cite{Kingma2013:VAE} architecture. However,  neural ODE models require deploying a numerical ODE solver which is computationally expensive. \citet{Bilos2021:neural-flow} hence propose an efficient alternative by directly modeling the solution of ODEs with a neural network, thereby obtaining a variant of Latent-ODE using Neural Flows, referred to as Latent-Flow in this paper. However, when analyzing time series, these Latent-based continuous-time models (Latent-ODE and Latent-Flow) require sequential processing of data, which makes them inefficient and hard to train.

In this work, we propose IVP-VAE, a continuous-time model specifically designed for EHR time series, which is capable of dealing with irregularly sampled time series data in a non-sequential way. Different from Latent-ODE and Latent-Flow, our model takes variational approximation purely as solving initial value problems (IVPs). Specifically, observations at different time points are mapped to states of an unknown continuous process and propagated to a latent variable $z_0$ by solving different IVPs in parallel. This parallelization leads to a significant speedup over existing continuous-time models. Latent-based continuous-time models use the VAE architecture, whose encoder and decoder consist of separate recognition and generative modules. We observe that neural IVP solvers are inherently invertible, i.e., IVPs can be solved in both forward and backward time directions, and exploit this property to utilize the same solver for both encoding and decoding. Our design results in reduced model complexity in terms of number of parameters and convergence rate.  

We deploy our model on the tasks of time series forecasting and classification across three real-world EHR datasets. IVP-VAE generally outperforms the existing latent-based continuous-time models across all the datasets and tasks. More importantly, it achieves more than one order of magnitude speedup over its latent-based predecessors. With regard to the state-of-the-art irregular sampled time series classification and forecasting models, IVP-VAE consistently ranks among the top-2 models, even though the baselines are in many cases task-specific. IVP-VAE offers the best performance trade-off across all the tasks. Additionally, our model is able to achieve significant improvements in settings where the training data is limited, which is often encountered in healthcare applications (e.g., cohort of patients with a particular condition).
We summarize the main contributions of the current work below -
\begin{itemize}
    \item We propose a novel continuous-time model IVP-VAE, which can capture sequential patterns of EHR time series by purely solving multiple IVPs in parallel.
    \item By utilizing the invertibility property of IVP solvers, we achieve parameter sharing between encoder and decoder of the VAE architecture, and thus provide a more efficient generative modeling technique.
    \item Across real-world datasets on both forecasting and classification tasks, IVP-VAE achieves a higher efficiency compared to the existing continuous-time models, and comparable performance to other state-of-the-art models. 
    \item IVP-VAE achieves significant improvements over baseline models in settings where the training data is limited.
\end{itemize}
Source code is at \url{https://github.com/jingge326/ivpvae}.

\section{Background and Related Work}

EHR data contains comprehensive information about patients' health conditions and has empowered the research on developing personalized medicine \cite{abul2019personalized_medicine}. The availability of several large EHR datasets, including MIMIC-III \cite{Johnson2016:mimic-iii}, MIMIC-IV \cite{johnson2023:mimic-iv}, and eICU \cite{pollard2019:eicu}, has facilitated the development of deep learning models for this domain. Specific tasks like time series forecasting and mortality prediction have been widely used to test models' capability in data modeling and representation learning \cite{harutyunyan2019multitask,mcdermott2021comprehensive,purushotham2018benchmarking,schirmer2022:rnn}. Functions built upon these can be used to support early warning of deterioration, identify patients at risk, diagnosis, etc. \cite{gao2020machine,Syed2021:mimic-review}. 
However, EHR time series are usually irregularly sampled \cite{Zhang2022:raindrop}, i.e., the time interval between consecutive observations is not fixed, and only some or no observations are available at each timestamp, making them sparse and of variable length \cite{weerakody2021review}.

There has been significant progress in developing models that are naturally able to handle irregularly sampled time series as the input \cite{Shukla2020:survey}. Several studies propose recurrent models that add decay mechanisms to model the irregularity in observations while training \cite{Cao2018:rnn,Kim2018:rnn,Li2019:rnn}. For example, GRU-D \cite{Che2018:gru-d} uses a temporal decay mechanism that is based on gated recurrent units (GRUs) and incorporates missing patterns. However, along with recurrent units comes the unstable gradient issue, and difficulties in long sequence modeling and parallelizing \cite{lipton2015critical_rnn}. Another group of work introduces attention mechanisms into models for irregular time series \cite{chien2021:attn,Horn2020:attn,Shukla2021:mTAN,tipirneni2022:attn}. For example, Raindrop~\cite{Zhang2022:raindrop} combines attention with graph neural networks to model irregularity. Owing to quadratic computation complexity and high memory usage, deploying these models to longer sequences becomes practically infeasible~\cite{zhou2021informer}. Convolutional models for irregular time series formulate the convolutional kernels as continuous functions \cite{fey2018:cnn,li2020:cnn,Romero2022:ckconv}, enabling them to handle sequences with arbitrary size and irregular sample intervals. However, when dealing with arbitrary length sequences, they usually need to first pad missing entries with specific values (such as zero) \cite{Romero2022:ckconv}, which can introduce irrelevant data and conceal important information.

Neural ODEs \cite{Chen2018:neural-ode} are continuous-time models that can naturally handle irregularly sampled data. The Latent ODE model \cite{Rubanova2019:latent-ode} uses an ODE-RNN encoder in a VAE \cite{Kingma2013:VAE} architecture. GRU-ODE-Bayes \cite{DeBrouwer2019:gru-ode-bayes} combines ODE and GRU into a continuous-time version of the GRU. Solving an ODE with a numerical ODE-Solver is computationally expensive.
Neural Flow \cite{Bilos2021:neural-flow} proposes an efficient alternative. The solution of an ODE is modelled directly with a neural network instead of using a numerical solver (see methodology section about continuous-time models for details).
A shortcoming of current research in this area is that existing methods often require sequentially solving a large amount of ODEs, which makes the training and inference less efficient. 

\subsubsection{Present Work}
Our method builds on VAE-based continuous models with Neural ODE and Neural Flow as IVP solvers. We introduce a set of novel architectural designs to further improve the efficiency. An embedding layer maps the input into a latent space where the IVP solvers are deployed. We eliminate the need for recurrent and sequential computation by modeling each time point as an IVP. As the IVP solvers are invertible by design, we propose to use the same IVP solver in the encoder and decoder of a VAE.

\section{Methodology}
In this section, we first formulate the problem, followed by a brief background on continuous-time models. We then introduce and describe our model in detail.

\subsection{Problem Formulation} \label{sect:prob_form}
In our setup, we consider a multivariate time series $X$ as a sequence of $L$ observations: $X = \{(\boldsymbol{x}_i, t_i)\}_{i=1}^L$. Each observation $\boldsymbol{x}_i$ is collected at a time $t_i$. $\boldsymbol{x}_i\in \mathbf{R}^D$ where $D$ represents the number of variables being measured at each time point (e.g., in EHR data these could represent a patient's heart rate, respiratory rate etc.). The dataset $\mathcal{X}$ consists of $N$ such sequences, $\mathcal{X} = \{X_1,\dots, X_N\}$, collected within a fixed time window. Note that the length $L$ of the sequences can vary across the dataset due to the irregular spacing of the observation time points.

Our goal is to first build a generative model $g$ for irregularly sampled time series (like EHR), which is capable of forecasting future values, and additionally augment it with a classifier to conduct classification tasks for which $g$ serves as a representation learning module. The time series forecasting task is to predict observations $X^{\tau}$ collected in time window $[T, T+\tau]$, based on past observations $X$, where $\tau$ is the forecast horizon. The classification task is to predict the categorical label $y$ of the sample $X$.

\begin{figure*}
\centering
\includegraphics[scale=0.75]{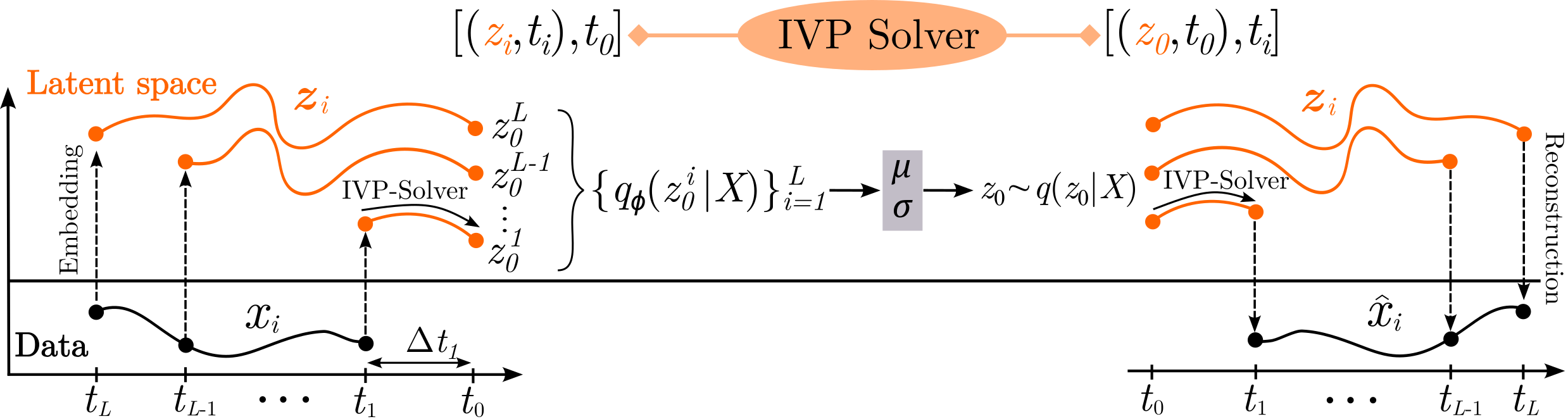}
\caption{Modeling irregular time series with IVP-VAE. (Left) In the encoder, an embedding module maps data $\boldsymbol{x}_i$ into latent state $\boldsymbol{z}_i$. The state is evolved backward in time: Take $(\boldsymbol{z}_i, t_i)$ as initial condition and calculate state $\boldsymbol{z}_0$ at $t_0$ using an IVP solver. (Right) In the decoder, the latent state is evolved forward in time: Take $(\boldsymbol{z}_0, t_0)$ as initial condition and go opposite along the timeline to obtain state $\boldsymbol{z}_i$ using the same IVP solver. A reconstruction module then maps $\boldsymbol{z}_i$ back to data $\boldsymbol{\hat{x}}_i$.}
\label{fig:ivpvae}
\end{figure*}

\subsection{Continuous-time Models}\label{sec:continuous-models}

A continuous-time model \cite{Chen2018:neural-ode} assumes that the data $\boldsymbol{x}_t$ at time $t$ is generated by a latent process $F$ whose state $\boldsymbol{z}$ can be propagated continuously to serve diverse purposes, such as generative modeling or representation learning. 
The propagation is achieved by solving IVPs, which are ODEs together with initial conditions, i.e.
\begin{align}
\label{eqn:ivp_solver}
& F\left(t_{0}\right)=\boldsymbol{z}_{0} \\
& f(t, \boldsymbol{z}_t)=\frac{d F(t)}{d t} \\
& \boldsymbol{z}_{i}=\boldsymbol{z}_{i-1}+\int_{t_{i-1}}^{t_{i}} f(t, \boldsymbol{z}_t) d t 
\end{align}
Neural ODEs \cite{Chen2018:neural-ode} parameterize $f$ with uniformly Lipschitz continuous neural networks, which are used to specify the derivative at every $t \in [t_0, T]$. States of the continuous process were calculated using the Runge–Kutta method or other numeric integrators. Neural Flow \cite{Bilos2021:neural-flow} proposes to directly model the solution curve $F$ with invertible neural networks. Both Neural ODEs and Neural Flow propagate hidden states by solving IVPs.

These continuous-time models are combined with recurrent units to analyze irregular time series. Given a hidden state $\boldsymbol{z}_{i-1}$, the idea to obtain the next hidden state is to propagate $\boldsymbol{z}$ using an IVP solver until the next observation $\boldsymbol{x}_{i}$ at time $t_{i}$ and then to use an RNN cell to update it, as expressed by the following equations.
\begin{align}
\label{eqn:ivp_rnn}
    & \boldsymbol{z}_{i}^{-} = \operatorname{IVPSolve}((\boldsymbol{z}_{i-1}, t_{i-1}), t_{i}) \\
    & {\boldsymbol{z}}_{i} = \operatorname{RNN}(x_i, \boldsymbol{z}_{i}^{-})
\end{align}

This represents the general IVP-RNN hybrid model first proposed by~\citet{Rubanova2019:latent-ode} using Neural ODEs as the IVP solver, known as Latent-ODE. The Latent-Flow variant of Latent-ODE uses Neural Flow as the IVP solver to directly obtain ${\boldsymbol{z}}_{i}^{-}$\cite{Bilos2021:neural-flow}. The entire model is trained as a VAE with the IVP-RNN hybrid model used as encoder to infer the posterior. Both Latent-ODE and Latent-Flow have proven to be effective in modeling irregular time series. However, the sequential nature of processing information makes these models computationally less efficient.

\subsection{Proposed Model: IVP-VAE}
The key idea that our model builds upon is that time series values $\left\{\boldsymbol{x}_{i}\right\}_{i=1}^{L}$ are discrete observations of an unknown continuous process. Each sample $X$ correspondingly represents a continuous process, of which we obtain an indirect observation at each available timestamp $t_i$. In this sense, our proposed model IVP-VAE is essentially a generative model for these continuous processes. From this idea, we design the model following two basic points: 
(i) We can circumvent the sequential operation bottleneck by processing all time steps independently as one ODE's different IVPs which can be solved in parallel. (ii) IVP solvers are inherently invertible, which enables us to use the same solver for both forward and backward propagation. The model is trained as a VAE 
whose encoder includes an embedding module and the IVP solver evolving latent state $\boldsymbol{z}_i$ backward in time, while the decoder includes the same IVP solver evolving the state forward in time, and a reconstruction module generating estimated data $\boldsymbol{\hat{x}}_i$ based on state $\boldsymbol{z}_i$.
The model is illustrated in Figure~\ref{fig:ivpvae} and the whole idea is summarized in Algorithm~\ref{alg:algorithm}.
The steps are described in the following sections. The IVP-VAE model can then be used for different downstream tasks, for example by appending a classification module.

\begin{algorithm}[t]
    \caption{IVP-VAE. The same IVP solver works for both encoder and decoder by solving IVPs in opposite directions.}
    \label{alg:algorithm}
    \textbf{Input}: Data points and timestamps $X=\left\{(\boldsymbol{x}_{i}, t_{i})\right\}_{i=1}^{L}$ \\
    \textbf{Output}: Reconstructed $\{\hat{\boldsymbol{x}}_i\}_{i=1}^L$ 
    \begin{algorithmic}[1] 
        \STATE $t_0 = 0$
        \STATE $\{\boldsymbol{z}_i\}_{i=1}^L = \operatorname{Embedding}(\{\boldsymbol{x}_i\}_{i=1}^L)$
        \STATE $\{\Delta t_i\}_{i=1}^L = t_0 - \{t_i\}_{i=1}^L$
        \STATE $\{\boldsymbol{z}^i_0\}_{i=1}^L = \{\operatorname{IVPSolve}\left(\boldsymbol{z}_i, \Delta t_i\right)\}_{i=1}^L$ \tikzmark{right}
        \STATE $q(\boldsymbol{z}_0|X) = \operatorname{Inference}\left(\{\boldsymbol{z}^i_0\}_{i=1}^L\right)$
        \STATE $\boldsymbol{z}_0 \sim q(\boldsymbol{z}_0|X)$
        \STATE $\{\Delta t_i\}_{i=1}^L = \{t_i\}_{i=1}^L - t_0$ \tikzmark{top}
        \STATE $\{\boldsymbol{z}_{i}\}_{i=1}^{L} = \operatorname{IVPSolve}\left(\boldsymbol{z}_0, \Delta t_i\right)\}_{i=1}^L$ \tikzmark{bottom}
        \rlap{\smash{$\left.\begin{array}{@{}c@{}}\\{}\\{}\end{array}\right\}\begin{tabular}{l}$p_{\theta}(X \mid z_0)$.\end{tabular}$}}
        \STATE $\{\hat{\boldsymbol{x}}_i\}_{i=1}^L = \operatorname{Reconstruct}(\{\boldsymbol{z}_i\}_{i=1}^L)$
        \STATE \textbf{return} $\{\hat{\boldsymbol{x}}_i\}_{i=1}^L$
    \end{algorithmic}
\end{algorithm}

\subsubsection{Embedding and Reconstruction}\label{sec:emb_rec}
Within the embedding module, given a time series $X=\left\{(\boldsymbol{x}_{i}, t_{i})\right\}_{i=1}^{L}$, we first generate corresponding binary masks $\left\{\boldsymbol{m}_{i}\right\}_{i=1}^{L}$ that indicate which variables are observed and which are not at time $t_i$. Next, we obtain $\boldsymbol{v}_i = (\boldsymbol{x}_{i}|\boldsymbol{m}_{i})$
for all observations at $t_i$ by concatenating $\boldsymbol{x}_{i}$ with $\boldsymbol{m}_{i}$. A neural network $\epsilon$ is then deployed on $\boldsymbol{v}_i$, $\boldsymbol{z}_i = \epsilon(\boldsymbol{v}_i)$, to extract useful information from multivariate observations at each timestamp, and produce $\boldsymbol{z}_i$ which represents the state of the continuous process at $t_i$.

On the decoder side, we design a similar module for data reconstruction that maps $\boldsymbol{z}_i$ to $\boldsymbol{x}_i$. The aim of adding embedding and reconstruction modules is to create a space in which the latent state $\boldsymbol{z}$ evolves, and also re-organize information into a more compact form. For these two modules, we use MLPs for demonstration and brevity. They can be more complex or well-designed networks. The embedding and the reconstruction operation are represented by line 2 and line 9 in Algorithm~\ref{alg:algorithm}, respectively.

\subsubsection{Evolving Backward in Time} \label{subsubsect_evolve_back}
Given that the true posterior $p(\boldsymbol{z}_{0} \mid X)$ is intractable \cite{Kingma2013:VAE}, the overall goal is to approximate the posterior, i.e., learn a variational approximation $q_{\phi}(\boldsymbol{z}_{0} \mid X)$ which can then be used to sample $z_0$.
For $\boldsymbol{x}_i$, the initial condition is defined as $(\boldsymbol{z}_i, t_i)$ in the encoder. The task of a neural IVP solver is to start from $t_i$, move towards $t_0$ continuously and calculate $\boldsymbol{z}_0$:
\begin{equation}
\label{eqn:ivp_encoder}
\boldsymbol{z}^{i}_{0} = \text{IVPSolve}(\boldsymbol{z}_{i}, \Delta t_i),
\end{equation}
where $\Delta t_i=t_0 - t_i$. $(\boldsymbol{z}_i, t_i)$ and $(\boldsymbol{z}_0^i, t_0)$ are on the same integral curve and satisfy the same ODE. Similarly, $\{\boldsymbol{z}^{i}_{0}\}_{i=1}^{L}$ is obtained for all $\{\boldsymbol{x}_{i}, \Delta t_{i} \}_{i=1}^{L}$. As we take observation $\boldsymbol{x}_i$ as an indirect observation of the unknown continuous process, we can make a guess of this process based on each $\boldsymbol{x}_i$, and then derive $\boldsymbol{z}^{i}_{0}$ of the process. Here, $\boldsymbol{z}^{i}_{0}$ is an estimation of $\boldsymbol{z}_{0}$ made by the IVP solver based on $\boldsymbol{x}_i$. Afterward, there are two issues to be addressed. First, each $\boldsymbol{z}^{i}_{0}$ should approximate $\boldsymbol{z}_{0}$ during training. Second, all $L$ $\boldsymbol{z}^{i}_{0}$ should be integrated together for the following generative module (decoder). For the first issue, we will discuss more details below in the training section. For the second issue, we define $q_{\phi}(\boldsymbol{z}_{0} \mid X)$ to be the posterior distribution over the latent variable $\boldsymbol{z}_0$ induced by the input time series $X$. To obtain it from $\left\{q_{\phi}(\boldsymbol{z}^{i}_{0} \mid X)\right\}_{i=1}^{L}$ in $\operatorname{Inference}$ (line 5 in Algorithm~\ref{alg:algorithm}), we introduce a mixture distribution over $\{\boldsymbol{z}^{i}_{0}\}_{i=1}^{L}$, constructed by diagonal Gaussian distribution $\mathcal{N}$
\begin{align}
    & q_{\phi}(\boldsymbol{z}^{i}_{0} \mid X) = \mathcal{N}(\boldsymbol{\mu}_{\boldsymbol{z}^{i}_{0}}, \boldsymbol{\sigma}_{\boldsymbol{z}^{i}_{0}}) \\
    & q_{\phi}(\boldsymbol{z}_{0} \mid X) = \sum_{i=1}^{L} \pi_i * q_{\phi}(\boldsymbol{z}^{i}_{0} \mid X) \label{eq_q_phi},
\end{align}
where $\boldsymbol{\mu}_{\boldsymbol{z}^{i}_{0}}=h(\boldsymbol{z}^{i}_0)$, and $\boldsymbol{\sigma}_{\boldsymbol{z}^{i}_{0}}=\operatorname{Softplus}(h(\boldsymbol{z}^{i}_0))$. $h$ denotes a feed-forward neural network and $\operatorname{Softplus}$ is the activation function. $\pi$ denote the mixing coefficients for the $L$ components. The entire operation is summarized by lines 3–5 in Algorithm~\ref{alg:algorithm}. More details about $\pi$ will be discussed below in the section about supervised learning.

\subsubsection{Evolving Forward in Time}
In this part, we first draw an instance from the posterior distribution $q_\phi(\boldsymbol{z}_{0} \mid X)$ to obtain 
$\boldsymbol{z}_{0}$ (line 6 in Algorithm~\ref{alg:algorithm}), which will further be used as a representation of the time series sample and also as the initial point of extrapolation.
We then start from $\boldsymbol{z}_0$ and propagate the latent state $\boldsymbol{z}$ forward along the timeline, with $\Delta t_i=t_i - t_0$ (line 7). Thus, $\boldsymbol{z}_{1}, \boldsymbol{z}_{2}, ..., \boldsymbol{z}_{L}$ can be calculated for all the $L$ timestamps by another call of the IVP solver (line 8):
\begin{equation}
    \{\boldsymbol{z}_{i}\}_{i=1}^{L} = \{\operatorname{IVPSolve}\left(\boldsymbol{z}_0, \Delta t_i\right)\}_{i=1}^L.
\end{equation}

The multivariate observation $\boldsymbol{x}_{i}$ can then be obtained from $\boldsymbol{z}_{i}$ using the data reconstruction module explained previously (line 9 in Algorithm~\ref{alg:algorithm}). The entire operation is mathematically represented as approximating $p_{\theta}(X \mid z_0 )$.

In terms of capturing temporal dependencies, RNN-related models repeatedly operate on sequential observations and extract useful information in an autoregressive way. Using IVP-VAE, the dependence is captured by Neural ODEs with derivatives, and Neural Flows with invertible transformations. Thus, the encoder and decoder do not require any recurrent operation, as all latent states at different time points can evolve independently given an ODE.

\subsubsection{Invertibility and Bidirectional Evolving}
The mechanism that one neural IVP solver works for both the encoder and decoder by solving IVPs in opposite time directions is achieved by utilizing the invertibility of IVP solvers. A detailed introduction to neural IVP solvers and the invertibility phenomena can be found in Appendix A.1. 

\subsection{Training} \label{subsect_training}
The IVP-VAE model can be trained both on unsupervised and supervised learning.
\subsubsection{Unsupervised Learning}
To learn the parameters of our IVP-VAE model given a dataset of sparse and irregularly sampled time series, we define the learning objective for one sample $X$ as
\begin{equation}
 \begin{aligned}
\mathcal{L}_{\operatorname{VAE}}(\phi, \theta)= &\mathbb{E}_{\boldsymbol{z}_0 \sim q_\phi(\boldsymbol{z}_0 \mid X)}\left[\log p_\theta\left(X \mid \boldsymbol{z}_0\right)\right] \\
& - \frac{1}{L} \sum_{i=1}^{L} D_{KL}(q_\phi(\boldsymbol{z}^i_0 \mid X) \| p(\boldsymbol{z}_0)),
\end{aligned}
\end{equation}
which corresponds to the evidence lower bound (ELBO) \cite{Kingma2013:VAE}.

As mentioned earlier, each $\boldsymbol{z}^{i}_{0}$ should approximate $\boldsymbol{z}_{0}$ during training, so the second term of $\mathcal{L}_{\operatorname{VAE}}(\phi, \theta)$ is the average of KL-divergence loss between $\left\{q_{\phi}(\boldsymbol{z}^{i}_{0} \mid X)\right\}_{i=1}^{L}$ and $p(\boldsymbol{z}_0)$. Given that not all data dimensions are observed at all time points, we calculate the reconstruction loss based on all available observations.

\subsubsection{Supervised Learning} \label{subsubsect_supervised}
\noindent\textbf{Forecasting.}
The model's capability of extrapolation can be used for time series forecasting. To produce value predictions out of the input time window $T$, one can simply continue to propagate the latent state $\boldsymbol{z}_i$ using the same neural IVP solver to any desired time points, e.g. in the forecast time window $[T, T+\tau]$, without adding any additional component. After propagation, the same reconstruction module can be used to map $\boldsymbol{z}_i$ to $\hat{\boldsymbol{x}}_i$, thus obtaining $\hat{X}^{\tau}$, which is the forecasted content with regard to the truth $X^{\tau}$. We combine $\mathcal{L}_{\operatorname{VAE}}$ with the reconstruction error $\mathcal{L}_{\operatorname{Re}}$ on $X^{\tau}$ to obtain Equation \ref{eqn:loss_forecast}, where $\alpha$ is a hyperparameter.
\begin{equation}
\label{eqn:loss_forecast}
\mathcal{L}_{\operatorname{Forecast}}(\phi, \theta)= \mathcal{L}_{\operatorname{VAE}}(\phi, \theta) + \alpha \cdot \mathcal{L}_{\operatorname{Re}}(\hat{X}^{\tau}\|X^{\tau})
\end{equation}

\noindent\textbf{Classification.}
We can also augment IVP-VAE with a classifier that leverages the latent state evolving as feature extraction and representation learning. We define this portable classification component to be of the form $p_\lambda\left(y \mid \boldsymbol{z}_0\right)$, where $\lambda$ represents model parameters (essentially a feed-forward network). This leads to an augmented learning objective, as shown in Equation \ref{eqn:loss_class}, where CE is the cross entropy loss. 
\begin{equation}
\begin{aligned}
\mathcal{L}_{\operatorname{Class}}(\phi, \theta, \lambda)= & \mathcal{L}_{\operatorname{VAE}}(\phi, \theta) \\
& + \alpha \cdot \operatorname{CE}(p(y)\|p_\lambda\left(y \mid \boldsymbol{z}_0\right))
\label{eqn:loss_class}
\end{aligned}
\end{equation}

The value of the mixing coefficient $\pi_i$ in Equation \ref{eq_q_phi} depends on the performed task. 
There exist various methods to determine the mixing coefficients in a mixture distribution. In our proposed model, we empirically obtained two different settings for the mixing coefficients: $\pi_i = \frac{1}{L}$ for classification and $\pi_i = \frac{D_{KL}(q_\phi(\boldsymbol{z}^i_0 \mid X) | p(\boldsymbol{z}_0))}{\sum_{j=1}^{L} D_{KL}(q_\phi(\boldsymbol{z}^j_0 \mid X) | p(\boldsymbol{z}_0))}$ for forecasting tasks.

\section{Experiments}
In this section, we present the experimental protocol and the  range of baseline models used along with the EHR datasets.

\subsection{Datasets }

We evaluate our model on three real-world public EHR datasets from the PhysioNet platform \cite{Goldberger2000:physionet}: MIMIC-IV~\cite{johnson2020:mimic-iv, johnson2023:mimic-iv}, PhysioNet 2012 \cite{Silva2012:Physionet2012} and eICU \cite{pollard2018eicu:eicu, pollard2019:eicu}.

The \textbf{MIMIC-IV} dataset is a multivariate EHR time series dataset consisting of sparse and irregularly sampled physiological signals collected at Beth Israel Deaconess Medical Center from 2008 to 2019. After data preprocessing following a similar procedure to \citet{Bilos2021:neural-flow}, 96 variables covering patient in- and outputs, laboratory measurements, and prescribed medications, are extracted over the first 48 hours after ICU admission. We obtain 26,070 records and use them for both forecasting and classification.

The \textbf{PhysioNet 2012} dataset was published as part of the PhysioNet/Computing in Cardiology Challenge 2012 with the objective of in-hospital mortality prediction. It includes vital signs, laboratory results, and demographics of patients admitted to an ICU. We use the provided 4,000 admissions from the challenge training set and 37 features over the first 48 hours after patient admission following \citet{Bilos2021:neural-flow}.

The \textbf{eICU} Collaborative Research Database is a multicenter dataset of patients admitted to ICUs at 208 hospitals located throughout the United States between 2014 and 2015. We follow the preprocessing procedure presented in \citet{Romero2022:ckconv} and extract 14 features over the first 48 hours after ICU admission for 12,312 admissions.

The key information of the three datasets after preprocessing is summarized in Table \ref{tab:datasets}. MIMIC-IV has the highest rate of missing values, the longest average sequence length, and the smallest positive rate for mortality. The eICU data is the least sparse, with a missing rate of only about 65\,\%. The full list of selected variables of each dataset can be found in Appendix A.3.

\begin{table}
\centering
\begingroup
\small
\begin{tabular}{l
ccc
ccc}
    \toprule
     & MIMIC-IV & PhysioNet 2012 & eICU\\
     \midrule
     \# Samples & 26,070 & 3,989 & 12,312 \\ 
     \# Variables & 96 & 37 & 14 \\ 
     Missing rate (\%) & 97.95 & 84.34 & 65.25 \\
     Average length & 173.4 & 75.0 & 114.55 \\
     Positive rate (\%) & 13.39 & 13.89 & 17.61 \\
     Granularity & 1 min & 1 min & 1 min \\
    \bottomrule
\end{tabular}
\endgroup
\caption[skip=50pt]{Key information of the three datasets after preprocessing: Number of admissions used, number of selected variables, overall percentage of missing values, average sequence length over admissions, positive rate for mortality, granularity of measurements.}
\label{tab:datasets}
\end{table}

\subsection{Baselines}
We compare our model against several baselines for the forecasting and classification of multivariate irregular time-series.

\begin{itemize}[leftmargin=*]
\setlength\itemsep{0em}
\item \textbf{GRU-$\Delta_{t}$} concatenates feature values with masking variable and time interval $\Delta_{t}$ as input \cite{Rubanova2019:latent-ode}.
\item \textbf{GRU-D} incorporates missing patterns using GRU combined with a learnable decay mechanism on both the input sequence and hidden states \cite{Che2018:gru-d}.
\item \textbf{mTAN} leverages an attention mechanism to learn temporal similarity and time embeddings \cite{Shukla2021:mTAN}.
\item \textbf{GRU-ODE-Bayes} couples continuous-time ODE dynamics with discrete Bayesian update steps \cite{DeBrouwer2019:gru-ode-bayes}.
\item \textbf{CRU} constructs continuous recurrent cells using linear stochastic differential equations and Kalman filters \cite{schirmer2022:rnn}.
\item \textbf{Raindrop} represents dependencies among multivariates with a graph whose connectivity is learned from time series \cite{Zhang2022:raindrop}.
\item \textbf{Latent-ODE} uses an ODE-RNN encoder and Neural ODE decoder in a VAE architecture \cite{Rubanova2019:latent-ode}.
\item \textbf{Latent-Flow} replaces the ODE component of Latent-ODE with more efficient Neural Flow models \cite{Bilos2021:neural-flow}.
\end{itemize}

Corresponding to two Latent-based models, we evaluate IVP-VAE with two types of IVP solvers, i.e. one with ODE called IVP-VAE-ODE and another with Flow called IVP-VAE-Flow. Hyperparameter settings are described in Appendix A.2.
Latent-ODE and Latent-Flow, which are the primary baselines for our model, are jointly referred to as \emph{Latent-based} models below.

\subsection{Experimental Protocols} \label{sect_exp_protocols}
All three datasets are used for forecasting and classification experiments. Each dataset is randomly split into 80\% for training, 10\% for validation and 10\% for testing. Following previous works \cite{Rubanova2019:latent-ode,Shukla2021:mTAN,Zhang2022:raindrop}, we repeat each experiment five times using different random seeds to split datasets and initialize model parameters.

In forecasting experiments, we use the first 24 hours of data as input and prediction the next 24 hours of data. We assess models' performance using the mean squared error (MSE). For classification experiments, we focus on predicting in-hospital mortality using the first 24 hours of data. Due to class imbalance in these datasets, we assess classification performance using area under the ROC curve (AUROC) and area under the precision-recall curve (AUPRC). To compare the models' running speed, we also report T-epoch \cite{Bilos2021:neural-flow,Horn2020:attn,Shukla2021:mTAN}, which is the time that each model needs to complete one epoch (counted in seconds). All models were tested in the same computing environment with NVIDIA Tesla V100 GPUs.

Considering the fact that even though some public EHR datasets have sufficient general samples for training complex deep learning models, when it comes to a specific group of patients or a specific medical phenomenon, the available data for training are usually not sufficient \cite{shickel2017deep}. We also deploy our model and other baselines in experiments with limited samples, conducting a comprehensive comparison across various dataset sizes. The samples are drawn from the MIMIC IV dataset. The test dataset consistently contains 2,000 samples, whereas the number of samples for training and validation ranges from 250 to 4,000. These small-sized datasets are then divided into training and validation sets at a 4:1 ratio. Both classification and forecasting tasks are conducted within this setting.

\section{Results and Analyses}

In this section, we evaluate IVP-VAE's capability of data modeling and representation learning for EHR time series data.
There are different branches of methods for irregular time series. We first make a thorough comparison of our designs and their Latent-based predecessors to show the improvement. Afterward, we compare our designs against the state-of-art and representative methods from other branches.

\begin{table}[!ht]
\centering
\begingroup
\small
\setlength{\tabcolsep}{4pt} 
\begin{tabular}{llrrrrrr}
    \toprule
     & & \multicolumn{2}{c}{ODE} & \multicolumn{2}{c}{Flow} \\
     \cmidrule(lr){3-4} \cmidrule(lr){5-6}
     & & \multicolumn{1}{c}{IVP-VAE} & \multicolumn{1}{c}{Latent} & \multicolumn{1}{c}{IVP-VAE} & \multicolumn{1}{c}{Latent} \\
     \midrule
     \multicolumn{2}{c}{MIMIC-IV} \\
     \cmidrule(lr){1-2}
     \multirow{6}{*}{\rotatebox[origin=c]{90}{Classification}}
     & AUROC & \textbf{0.802} & 0.768 & \textbf{0.805} & 0.786 \\ 
     & AUPRC & \textbf{0.422} & 0.393 & \textbf{0.427} & 0.404 \\ 
     & T-forward & \textbf{0.066} & 2.536 & \textbf{0.017} & 0.784 \\ 
     & T-epoch & \textbf{1478.8} & 5270.3 & \textbf{1445.8} & 3105.5 \\
     & \# Epochs & \textbf{12.6} & 56.2 & \textbf{10.8} & 57.8 \\
     & \# Parameters & 209,677 & \textbf{199,017} & \textbf{325,017} & 429,697 \\ [0.3mm]
     \hdashline \\ [-2.8mm]
     \multirow{5}{*}{\rotatebox[origin=c]{90}{Forecasting}}
     & MSE & \textbf{0.724} & 0.769 & \textbf{0.727} & 0.755 \\ 
     & T-forward & \textbf{0.106} & 3.059 & \textbf{0.025} & 1.063 \\ 
     & T-epoch & \textbf{155.4} & 4294.9 & \textbf{81.5} & 2272.0 \\
     & \# Epochs & \textbf{31.8} & 37.2 & \textbf{35.6} & 42.8 \\
     & \# Parameters & 112,776 & \textbf{102,116} & \textbf{228,116} & 332,796 \\
     \midrule
     \multicolumn{2}{c}{PhysioNet 2012} \\
     \cmidrule(lr){1-2}
     \multirow{6}{*}{\rotatebox[origin=c]{90}{Classification}}
     & AUROC & \textbf{0.770} & 0.767 & \textbf{0.771} & 0.766 \\ 
     & AUPRC & 0.359 & \textbf{0.364} & \textbf{0.362} & 0.327 \\ 
     & T-forward & \textbf{0.031} & 1.072 & \textbf{0.009} & 0.285 \\ 
     & T-epoch & \textbf{35.6} & 333.4 & \textbf{32.6} & 166.5 \\
     & \# Epochs & \textbf{19.6} & 89.2 & \textbf{19.4} & 96.2 \\
     & \# Parameters & \textbf{174,218} & 188,338 & \textbf{289,558} & 419,018 \\ [0.3mm]
     \hdashline \\ [-2.8mm]
     \multirow{5}{*}{\rotatebox[origin=c]{90}{Forecasting}}
     & MSE & \textbf{0.563} & 0.586 & \textbf{0.567} & 0.584 \\ 
     & T-forward & \textbf{0.072} & 0.916 & \textbf{0.012} & 0.307 \\ 
     & T-epoch & \textbf{20.2} & 292.9 & \textbf{8.2} & 264.7 \\
     & \# Epochs & 54.4 & \textbf{38.4} & 68.0 & \textbf{44.2} \\
     & \# Parameters & \textbf{77,317} & 91,437 & \textbf{192,657} & 322,117 \\
     \midrule
     \multicolumn{2}{c}{eICU} \\
     \cmidrule(lr){1-2}
     \multirow{6}{*}{\rotatebox[origin=c]{90}{Classification}}
     & AUROC & \textbf{0.786} & 0.783 & \textbf{0.786} & 0.781 \\ 
     & AUPRC & 0.468 & \textbf{0.477} & 0.472 & \textbf{0.482} \\ 
     & T-forward & \textbf{0.033} & 2.733 & \textbf{0.009} & 0.539 \\ 
     & T-epoch & \textbf{342.5} & 2296.1 & \textbf{319.4} & 1127.4 \\
     & \# Epochs & \textbf{16.0} & 78.8 & \textbf{23.0} & 93.0 \\
     & \# Parameters & \textbf{160,395} & 184,175 & \textbf{275,735} & 414,855 \\ [0.3mm]
     \hdashline \\ [-2.8mm]
     \multirow{5}{*}{\rotatebox[origin=c]{90}{Forecasting}}
     & MSE & \textbf{0.596} & 0.598 & \textbf{0.581} & 0.594 \\ 
     & T-forward & \textbf{0.081} & 2.604 & \textbf{0.012} & 0.655 \\ 
     & T-epoch & \textbf{77.7} & 1778.3 & \textbf{28.2} & 616.3 \\
     & \# Epochs & 60.2 & \textbf{32.0} & 78.2 & \textbf{42.5} \\
     & \# Parameters & \textbf{160,395} & 184,175 & \textbf{275,735} & 414,855 \\
    \bottomrule
\end{tabular}
\endgroup
\caption[skip=50pt]{Detailed comparison of IVP-VAE and its predecessor using different IVP solvers on three datasets for classification and forecasting. We compare the time needed for one forward pass (T-forward) and for one epoch (T-epoch), number of epochs, and number of parameters. Better results are in bold.}
\label{tab:exp_compare_vaes}
\end{table}

\begin{table*}[t]
\centering
\begingroup
\small
\setlength{\tabcolsep}{4pt} 
\begin{tabular}{l
ccc
ccc
ccc
}
    \toprule
     & \multicolumn{3}{c}{MIMIC-IV} & \multicolumn{3}{c}{PhysioNet 2012} & \multicolumn{3}{c}{eICU} \\ 
    \cmidrule(lr){2-4} \cmidrule(lr){4-7} \cmidrule(lr){6-10} 
     & MSE & AUROC & AUPRC & MSE & AUROC & AUPRC & MSE & AUROC & AUPRC \\
    \midrule
     GRU-$\Delta_{t}$ & 0.730$ \pm $0.014 & \textbf{80.9$ \pm $0.6} & 42.0$ \pm $2.0 & 0.587$ \pm $0.055 & 72.0$ \pm $4.4 & 29.0$ \pm $4.5 & 0.583$ \pm $0.009 & 76.1$ \pm $1.4 & 42.8$ \pm $2.1 \\ 
     GRU-D & 0.736$ \pm $0.005 & 78.6$ \pm $0.9 & 41.9$ \pm $1.3 & 0.588$ \pm $0.060 & 76.2$ \pm $3.2 & 32.9$ \pm $4.3 & 0.578$ \pm $0.007 & \textbf{79.6$ \pm $1.5} & \textbf{47.7$ \pm $2.4} \\
     mTAN & 0.715$ \pm $0.011 & 76.6$ \pm $0.6 & 37.9$ \pm $2.4 & 0.588$ \pm $0.050 & 76.2$ \pm $2.8 & 33.8$ \pm $5.3 & 0.582$ \pm $0.010 & 76.9$ \pm $2.4 & 45.1$ \pm $3.2 \\ 
     Raindrop & - & 77.1$ \pm $1.4 & 36.8$ \pm $2.8 & - & 75.3$ \pm $2.3 & 30.9$ \pm $3.9 & - & 76.6$ \pm $2.1 & 45.1$ \pm $2.7 \\
     GOB & 0.809$ \pm $0.014 & - & - & 0.619$ \pm $0.029 & - & - & 0.664$ \pm $0.012 & - & - \\
     CRU & 0.946$ \pm $0.016 & - & - & 0.688$ \pm $0.032 & - & - & 0.820$ \pm $0.044 & - & - \\
     \hdashline \\ [-2.5mm]
     IVP-VAE-Flow & \textbf{0.727$ \pm $0.013} & 80.5$ \pm $0.5 & \textbf{42.7$ \pm $1.4} & \textbf{0.567$ \pm $0.038} & \textbf{77.1$ \pm $3.0} & \textbf{36.2$ \pm $5.3} & \textbf{0.581$ \pm $0.009} & 78.6$ \pm $1.7 & 47.2$ \pm $0.029 \\
    \bottomrule
\end{tabular}
\endgroup
\caption[skip=50pt]{Comparison of the proposed IVP-VAE-Flow model and state-of-the-art baselines. '-' denotes that a model doesn't support the task. We report test MSE for forecasting and AUROC $\left(\times 10^{-2}\right)$ and AUPRC $\left(\times 10^{-2}\right)$ for mortality prediction on three datasets. IVP-VAE-Flow achieves competitive performance across all datasets and tasks.}
\label{tab:combined_results}
\end{table*}

\subsection{Improvements Over Latent-Based Models}
To have a clear view of the improvement of performance and efficiency, we make a detailed comparison of our design and Latent-based models in Table \ref{tab:exp_compare_vaes}. Regarding performance in classification (AUROC \& AUPRC, larger means better) and forecasting (MSE, smaller means better) tasks, IVP-VAE generally outperforms Latent-based models across all datasets. 

We further compare efficiency of these models in terms of T-epoch and T-forward (the time taken by each model to complete one forward run). Clearly, IVP-VAE models are able to achieve a significant speed advantage over the corresponding Latent-based models. 
For instance, on forecasting tasks of MIMIC-IV, IVP-VAE-Flow is about 42 times faster than Latent-Flow in terms of T-forward. 
Since T-epoch includes T-forward as well as the time for data loading, loss calculation, backpropagation, etc., which significantly contribute to the computation time, the improvement in T-epoch is not as significant as in T-forward. Nevertheless, IVP-VAE-Flow is still more than 28 times faster than Latent-Flow. The speed advantage is achieved by eliminating recurrent operations and solving IVPs in parallel.

Furthermore, we compare IVP-VAE with its counterparts on convergence rate. As indicated by \# Epochs, IVP-VAE models converge significantly faster than Latent-based models, with IVP-VAE models needing lesser epochs to achieve the best validation accuracy. 
This advantage is achieved by the parameter sharing mechanism in our models, i.e. one IVP solver for both the encoder and decoder.
Multiplying the time per epoch by the number of epochs to obtain the total training time, we find that IVP-VAE based models are at least one order of magnitude faster than Latent-based models for both classification and forecasting tasks.

Regarding model size (\# Parameters in Table \ref{tab:exp_compare_vaes}), IVP-VAE-Flow is smaller than Latent-Flow in all scenarios. IVP-VAE-ODE is smaller than Latent-ODE in most cases, except for forecasting tasks on MIMIC-IV. Compared to Latent-based models, IVP-VAE (1) eliminates recurrent units, (2) uses one IVP solver instead of two, and (3) adds in the embedding and reconstruction modules. Factor (1) and (2) can reduce the number of parameters, while factor (3) increases the number of parameters. The overall parameter difference is the result of the superposition of these three factors.

\subsection{Comparison Against Other Representative Models}

We compare IVP-VAE-Flow as the best-performing proposed model with other state-of-the-art and representative baselines. The results of forecasting (in MSE) and classification (AUROC, AUPRC) experiments on the three datasets are presented in Table~\ref{tab:combined_results}. For each metric, we use bold font to indicate the best result. When compared with other state-of-the-art baseline models, the IVP-VAE-Flow model consistently achieves at least the second-best result. Also, IVP-VAE even achieves the best results for the PhysioNet 2012 dataset for both forecasting and classification. Overall, the proposed method exhibits competitive performance across all the datasets and tasks.

\subsection{Experiments on Small Datasets}

\begin{figure}[t]
\centering
\includegraphics[scale=0.245]{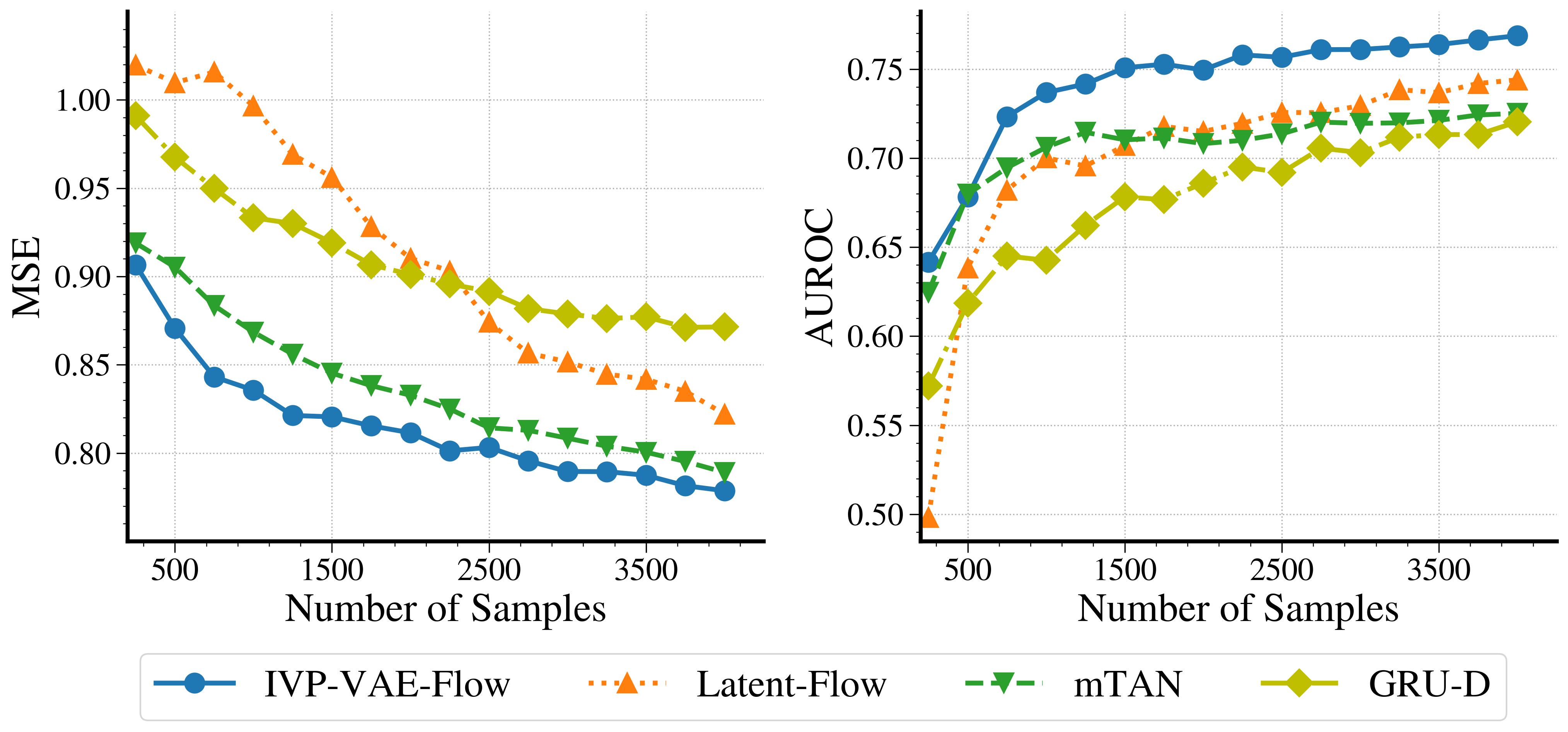}
\caption{Performance comparison on small datasets: (Left) MSE for forecasting and (right) AUROC for classification task.
IVP-VAE based models consistently and substantially outperform all baseline approaches across all datasets with different number of samples.}
\label{fig:curve_test_all}
\end{figure}

To further demonstrate the capabilities of the proposed model, we examine the performance under low sample size conditions.
This scenario is analogous to a rare disease setting in the field of EHR prediction, where data can only be obtained for a small cohort of patients. In such cases, the effectiveness of models in capturing temporal evolving patterns and rapidly updating parameters becomes essential. Figure~\ref{fig:curve_test_all} compares the performance of 4 typical methods on small datasets where we collected only a limited number of samples for model training and validation. As we can see, for both forecast and classification tasks, IVP-VAE based model consistently and substantially outperforms other approaches across all settings with different sample sizes. The advantage of the model on small datasets is also due to its parameter sharing mechanism in the encoder and decoder.

\section{Conclusion and Discussion}

In this paper, we have presented a faster and lighter continuous-time generative model IVP-VAE, which is able to model and learn representations of irregular sampled EHR time series by purely solving IVPs in parallel under the VAE architecture. Our results showed that the proposed models perform comparable or better than other baselines on classification and forecasting tasks, while offering training times that are one order of magnitude faster than previous continuous-time methods. Further experiments on small datasets showed that our model has an advantage in scenarios where the number of training samples is limited. Based on this, more work can be done to demonstrate the ability of IVP-VAE to model irregular sample time series with diverse datasets, not only EHR datasets, and different tasks like missing value imputation, time series regression, etc.

\section{Acknowledgments}
This research was funded by the Federal Ministry of Education and Research (BMBF), Germany under the project LeibnizKILabor with grant No. 01DD20003.

\bibliography{references}
\newpage

\section{Appendix}

\subsection{A.1 Neural IVP solvers and their invertibility} \label{appendix:ivps_invert}

\subsubsection{Initial Value Problem}

An initial value problem (IVP) is an ordinary differential equation together with an initial condition which specifies the value of the unknown function at a given point in the domain, i.e.
\begin{align}
F\left(t_{0}\right)=\boldsymbol{z}_{t_{0}} \\
\frac{d F(t)}{d t}=f(t, \boldsymbol{z}_t)
\end{align}
where the differential equation describes the rate of change of the function $F$ with respect to the variable $t$. $t_0$ is the given point in the domain, often means time in many problems. $F\left(t_{0}\right)=\boldsymbol{z}_{t_{0}}$ is the initial condition, which specifies the value of the function $F$ at the point $t_0$. An initial value problem aims to find the function $F(t)$ that satisfies both the differential equation and the initial condition. 

Neural ODEs parameterize $f$ with uniformly Lipschitz continuous neural networks, which are used to specify the derivative at every $t \in [t_0, T]$. States of the continuous process were calculated using the Runge–Kutta method or other numeric integrators. Neural Flow proposes to directly model the solution curve $F$ with invertible neural networks. As both of them can numerically solve initial value problems, we collectively refer to them as initial value problem solvers.

\subsubsection{Invertibility}

As for Neural ODEs, the state $\boldsymbol{z}$ at different timestamps can be computed according to 
$$\boldsymbol{z}_{i}=\boldsymbol{z}_{i-1}+\int_{t_{i-1}}^{t_{i}} f(t, \boldsymbol{z}_t) d t$$
here $f$ is determined within its domain of definition while $\Delta t$ can be negative or positive. If positive, the state evolves forward in time. If negative, it evolves backward in time. 

Similarly, a flow $\xi$ is mathematically defined as a group action on a set $Z$,
\begin{equation}
    \xi: Z \times \mathbb{R} \rightarrow Z,
\end{equation}
where for all $\boldsymbol{z} \in Z$ and real number $t$, 
\begin{align}
& \xi(\boldsymbol{z}, 0)=\xi(\boldsymbol{z}), \\
& \xi(\xi(\boldsymbol{z}, t_i), t_j)=\xi(\boldsymbol{z}, t_j+t_i).
\end{align}
This characteristic naturally guarantees its invertibility. And if $t_j = -t_i$, then 
\begin{equation}
    \xi(\xi(z, t_i), t_j)=\xi(z,0)=\xi(z)
\end{equation}
can describe the whole evolving process from encoder to decoder. Neural Flows can be trained within the VAE framework forward/backward in time to obtain this behavior.

\subsubsection{An Example}

\begin{figure}[ht]
\centering
\includegraphics[width=0.4\textwidth]{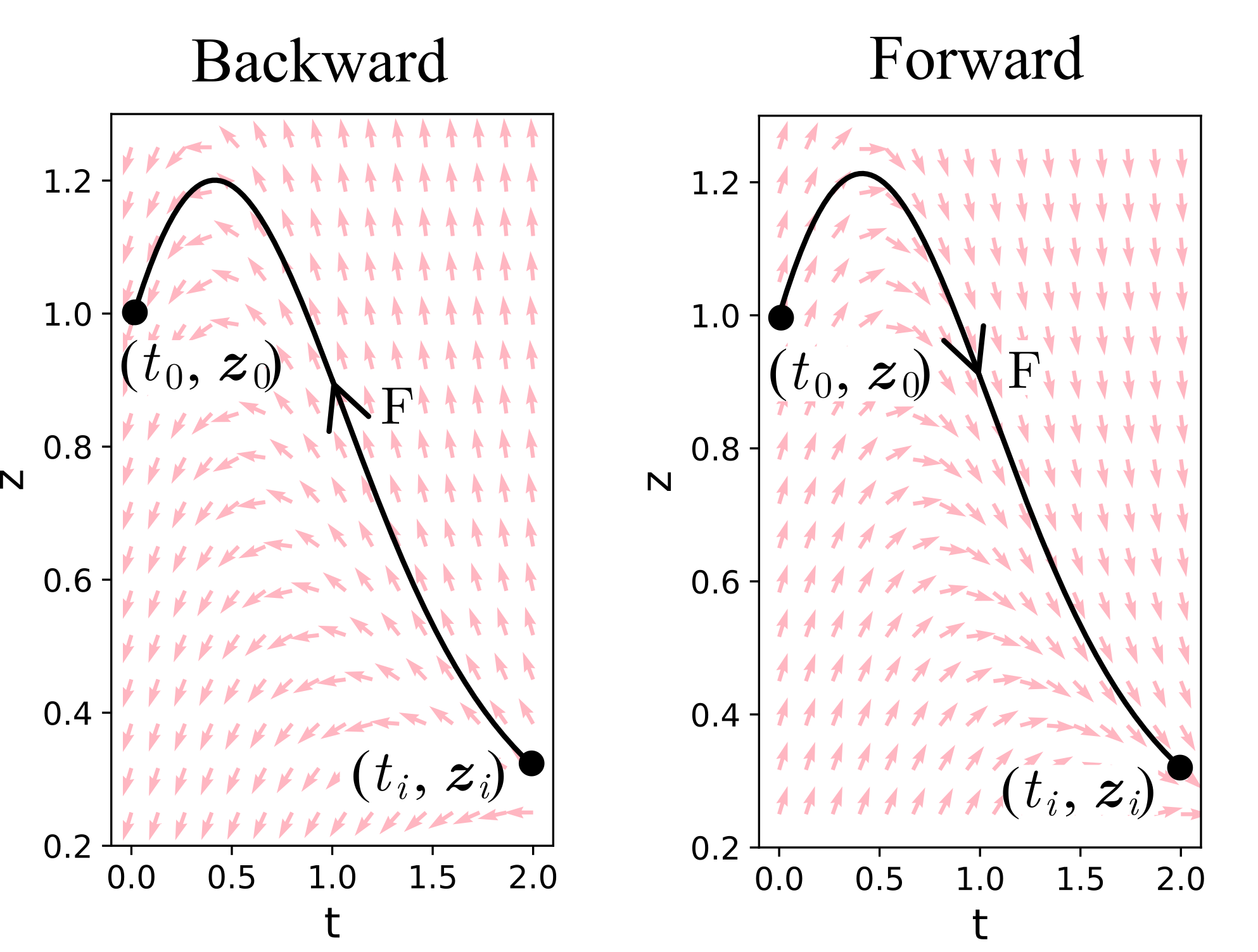}
\caption{Evolving forward and backward in time using the same IVP solver and different initial points}
\label{fig:ivp_backward_forward}
\end{figure}

Intuitively, in figure \ref{fig:ivp_backward_forward}, given an ODE 
$$\frac{d\boldsymbol{z}}{dt} = 1 - 2 \cdot t \cdot \boldsymbol{z}$$
If we take $(\boldsymbol{z}_i, t_i)$ as the initial condition (in the left sub-figure), where $\boldsymbol{z}=0.31953683$ and $t=2.0$, and calculate the state at $t=0.0$ using an IVP solver F, we can obtain $\boldsymbol{z}=1.0$. Meanwhile, in the right sub-figure, if taking $(\boldsymbol{z}=1.0, t=0.0)$ as the initial condition and evolve forward in time along the slope field using F, we can obtain $\boldsymbol{z}=0.31953683$. $(\boldsymbol{z}_i, t_i)$ and $(\boldsymbol{z}_0, t_0)$ are on the same integral curve, one can calculate forward and backward using the same IVP solver.

\subsection{A.2 Hyperparameter and detailed model settings}\label{appendix:hypers}

\noindent\textbf{All experiments} 
\begin{itemize}
    \item Optimizer: Adam
    \item Weight decay: 1e-4
    \item Batch size: 50
    \item Learning rate: 1e-3
    \item Learning rate scheduler step: 20
    \item Learning rate decay: 0.5
\end{itemize}

\noindent\textbf{Model detailed settings}

\noindent\textbf{IVP-VAE}
\begin{itemize}
    \item Latent state dimension: 20
    \item $\alpha$ balancing the cross-entropy loss and ELBO
    \begin{itemize}
        \item MIMIC-IV: 1000
        \item PhysioNet 2012: 100
        \item eICU: 100
    \end{itemize}
    \item $\alpha$ balancing forecasting loss and ELBO: 1
    \item Variants
    \begin{itemize}
        \item \textbf{IVP-VAE-Flow}
        \begin{itemize}
            \item Flow model: ResNet flow
            \item Number of flow layers: 2
        \end{itemize}
    \end{itemize}
    \begin{itemize}
        \item \textbf{IVP-VAE-ODE}
        \begin{itemize}
            \item Integrator: dopri5
        \end{itemize}
    \end{itemize}
\end{itemize}

\noindent\textbf{Latent-based models}
\begin{itemize}
    \item Latent state dimension: 20
    \item $\alpha$ balancing the cross-entropy loss and ELBO: 100
    \item $\alpha$ balancing forecasting loss and ELBO: 1
    \item Variants
    \begin{itemize}
        \item \textbf{Latent-Flow}
        \begin{itemize}
            \item Flow model: ResNet flow
            \item Number of flow layers: 2
        \end{itemize}
    \end{itemize}
    \begin{itemize}
        \item \textbf{Latent-ODE}
        \begin{itemize}
            \item Integrator: dopri5
        \end{itemize}
    \end{itemize}
\end{itemize}

\noindent\textbf{RNN-based models}
\begin{itemize}
    \item Hidden dimension: 20
    \item Variants
    \begin{itemize}
        \item \textbf{GRU-$\Delta_t$}
        \item \textbf{GRU-D}
    \end{itemize}
\end{itemize}

\noindent\textbf{mTAN}
\begin{itemize}
    \item Hidden dimension: 20
    \item Hidden dimension of the recognition module: 256
    \item Hidden dimension of the generative module: 50
    \item Dimension of the embedded time feature: 128
    \item Number of attention heads: 1
    \item Number of reference points: 128
    \item $\alpha$ balancing the cross-entropy loss and ELBO: 100
    \item $\alpha$ balancing forecasting loss and ELBO: 1
\end{itemize}

\noindent\textbf{Raindrop}
\begin{itemize}
    \item Dimension of the position encoder: 16
    \item Number of attention heads: 4
    \item Dropout rate: 0.2
    \item Number of Transformer Encoder layers: 2
    \item Time downsampling rate: 0.1
\end{itemize}

\noindent\textbf{CRU}
\begin{itemize}
    \item Number of hidden units: 50
    \item Number of basis matrices: 20
    \item Bandwidth for basis matrices: 10
\end{itemize}

\noindent\textbf{GRU-ODE-Bayes}
\begin{itemize}
    \item Ratio between KL and update loss: 0.0001
    \item Size of hidden state for covariates: 10
    \item Size of hidden state for initialization: 25
\end{itemize}

\subsection{A.3 Selected variables} \label{appendix:sel_vars}
For a summary of the variables selected for the MIMIC-IV, PhysioNet 2012 and eICU datasets, see the following Table~\ref{tab:sel_vars}.

\begin{table*}[!ht]
    \centering
    \caption{Selected variables for MIMIC-IV, PhysioNet 2012 and eICU dataset.}
    \label{tab:sel_vars}
    \begingroup
    \scriptsize
    \setlength{\tabcolsep}{4pt} 
    \begin{tabular}{llll}
        \toprule
        \multicolumn{2}{c}{MIMIC-IV} & PhysioNet 2012 & eICU \\
        \midrule
        Potassium Chloride & Magnesium Sulfate & Albumin & HR \\
        Calcium Gluconate & PO Intake & ALP & MAP \\
        Insulin - Glargine & Insulin - Regular & ALT & Invasive BP Diastolic \\
        LR & Furosemide (Lasix) & AST & Invasive BP Systolic \\
        OR Crystalloid Intake & OR Cell Saver Intake & Bilirubin & O2 Saturation \\
        Solution & Dextrose 5\% & BUN & Respiratory Rate \\
        Piggyback & Phenylephrine & Cholesterol & Temperature \\
        KCL (Bolus) & Albumin 5\% & Creatinine & Glucose \\
        PT & PTT & DiasABP & Fi02 \\
        Basophils & Eosinophils & FiO2 & pH \\
        Hematocrit & Hemoglobin & GCS & Glasgow Coma Score Total \\
        Lymphocytes & MCH & Glucose & GCS Eyes \\
        MCV & Monocytes & HCO3 & GCS Motor \\
        Neutrophils & RDW & HCT & GCS Verbal \\
        Red Blood Cells & White Blood Cells & HR & ~ \\
        Anion Gap & Chloride & K & ~ \\
        Creatinine & Magnesium & Lactate & ~ \\
        Phosphate & Potassium & Mg & ~ \\
        Urea Nitrogen & Base Excess & MAP & ~ \\
        Calculated Total CO2 & pCO2 & MechVent & ~ \\
        pO2 & Lactate & Na & ~ \\
        Platelet Count & pH & NIDiasABP & ~ \\
        Bicarbonate & Sodium & NIMAP & ~ \\
        Specific Gravity & Glucose & NISysABP & ~ \\
        Foley & Chest Tube 1 & PaCO2 & ~ \\
        OR Urine & Sodium Chloride 0.9\%  Flush Drug & PaO2 & ~ \\
        Potassium Chloride Drug & Magnesium Sulfate Drug & pH & ~ \\
        Acetaminophen Drug & Docusate Sodium Drug & Platelets & ~ \\
        Aspirin Drug & Insulin Drug & RespRate & ~ \\
        Metoprolol Tartrate Drug & Bisacodyl Drug & SaO2 & ~ \\
        Calcium, Total & Void & SysABP & ~ \\
        OR EBL & Emesis & Temp & ~ \\
        Pantoprazole Drug & Heparin Drug & TroponinI & ~ \\
        Lorazepam (Ativan) & Heparin Sodium & TroponinT & ~ \\
        Midazolam (Versed) & Alanine Aminotransferase (ALT) & Urine & ~ \\
        Alkaline Phosphatase & Asparate Aminotransferase (AST) & WBC & ~ \\
        Bilirubin, Total & Albumin & Weight & ~ \\
        Gastric Meds & GT Flush & ~ & ~ \\
        Norepinephrine & Pre-Admission & ~ & ~ \\
        D5W Drug & Metoprolol & ~ & ~ \\
        Packed Red Blood Cells & Sterile Water & ~ & ~ \\
        D5 1/2NS & Magnesium Sulfate (Bolus) & ~ & ~ \\
        Oral Gastric & Straight Cath & ~ & ~ \\
        K Phos & Morphine Sulfate & ~ & ~ \\
        Insulin - Humalog & Nitroglycerin & ~ & ~ \\
        TF Residual & Jackson Pratt 1 & ~ & ~ \\
        TF Residual Output & Nasogastric & ~ & ~ \\
        Stool & Fecal Bag & ~ & ~ \\
        \bottomrule
    \end{tabular}
    \endgroup
\end{table*}

\subsection{A.4 Additional Experiments}

\subsubsection{Experiments on non-EHR Irregular Time Series Datasets}

The effective and precise analysis of EHR data holds crucial importance in the fields of early intervention and personalized medicine. EHR time series data, owing to its irregular nature, presents challenges that common deep learning models struggle to address and hence our decision to work with such data. However, EHR dataset is not a design requirement for our model and it is equally effective on non-EHR data which we demonstrate through experiments on three non-EHR datasets.

\begin{figure}[ht]
\centering
\includegraphics[scale=0.25]{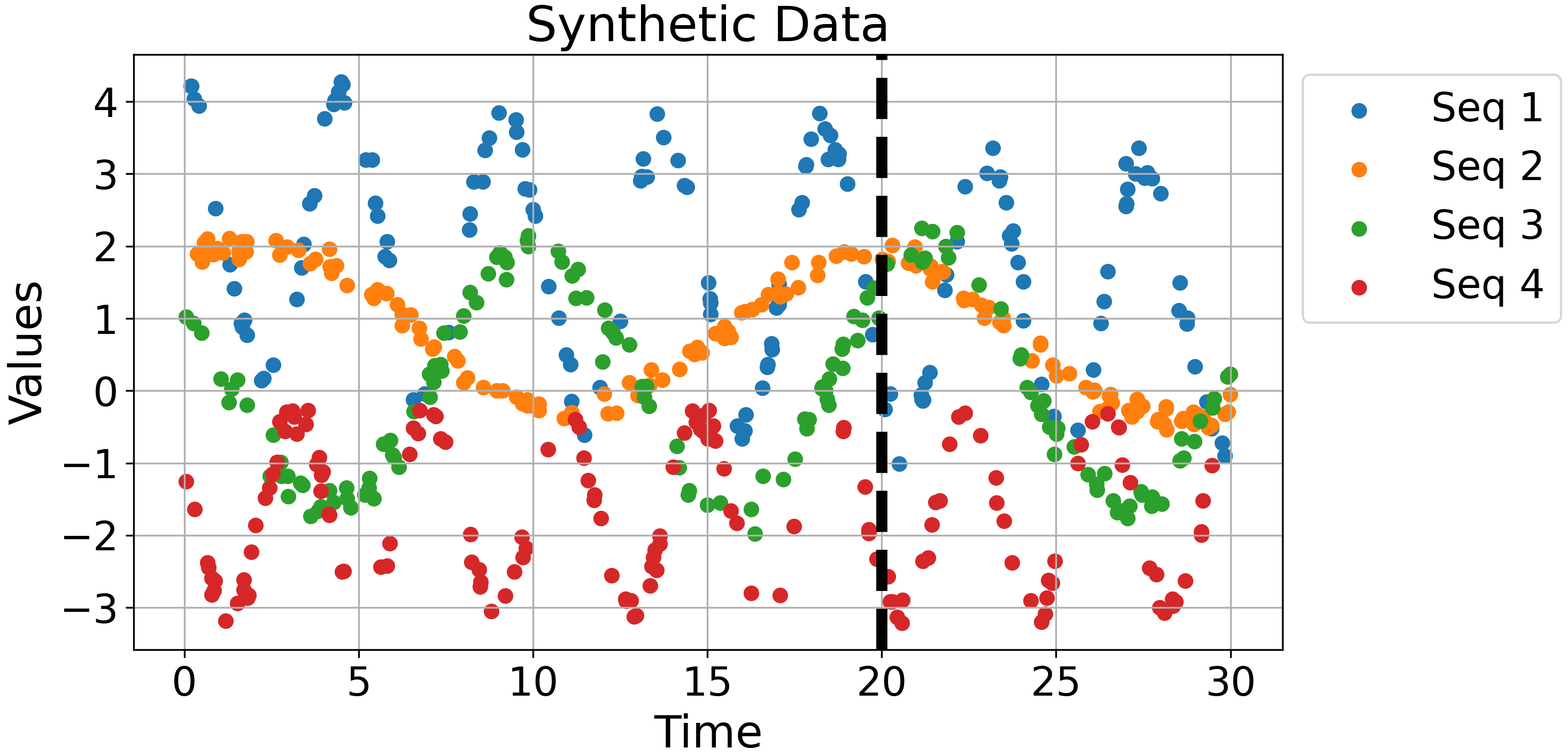}
\caption{Samples of the synthetic dataset}
\label{fig:synthetic}
\end{figure}

The first dataset is a synthetic dataset of 1,000 samples constructed following
$$
y {=} a {\cdot} sin(b {\cdot} t {+} c) {+} d {\cdot} t {+} e {+} noise
$$ 
$t {\in} [0, 30]$, $a {\in} [0.5, 2]$, $b {\in} [0.1, 2]$, $c {\in} [0, 2\pi]$, $d {\in} [-0.05, 0.05]$, $e {\in} [-4, 4]$ 
are randomly sampled from uniform distributions, and noise from a normal distribution ($\mu{=}0$, $\sigma{=}0.1$) is added.
Given values at $t {\in} [0, 20]$, models are trained to forecast values at $t {\in} [20, 30]$. Some samples of the synthetic dataset are shown in Figure~\ref{fig:synthetic}.
The second and third are two commonly used irregular time series datasets, Climate \cite{schirmer2022:rnn} for forecasting, and Activity \cite{Rubanova2019:latent-ode} for both forecasting and classification. Experiment results are shown in Table~\ref{tab:exp_irregular_ts}.
IVP-VAE models consistently rank among the best performing methods across all the datasets and tasks.

\begin{table}[ht]
\centering
\caption[skip=50pt]{Experiment on other irregular time series datasets}
\label{tab:exp_irregular_ts}
\begingroup
\scriptsize
\renewcommand{\arraystretch}{0.9}
\setlength{\tabcolsep}{3pt} 

\begin{tabular}{lcccc}
    \toprule
     & \multicolumn{1}{c}{Synthetic} & \multicolumn{1}{c}{Climate} & \multicolumn{2}{c}{Activity} \\
     \cmidrule(lr){2-2} \cmidrule(lr){3-3} \cmidrule(lr){4-5}
     & \multicolumn{1}{c}{MSE} & \multicolumn{1}{c}{MSE} & \multicolumn{1}{c}{MSE} & \multicolumn{1}{c}{Accuracy} \\
     \midrule
     GRU-$\Delta_{t}$ & 0.381$ \pm $0.055 & 0.594$ \pm $0.113 & 6.591$ \pm $0.153 & 0.750$ \pm $0.046 \\ 
     GRU-D & 0.385$ \pm $0.024 & 0.528$ \pm $0.061 & 6.483$ \pm $0.087 & \textbf{0.770$ \pm $0.081} \\ 
     mTAN & 0.222$ \pm $0.014 & \textbf{0.512$ \pm $0.094} & 6.284$ \pm $0.103 & 0.769$ \pm $0.025 \\ 
     Raindrop & - & - & - & 0.732$ \pm $0.101 \\
     GOB & 0.396$ \pm $0.054 & 0.634$ \pm $0.067 & 7.171$ \pm $0.140 & - \\
     CRU & 0.332$ \pm $0.047 & 0.517$ \pm $0.114 & 7.082$ \pm $0.118 & - \\
     Latent-ODE & 0.227$ \pm $0.038 & 0.541$ \pm $0.102 & 6.390$ \pm $0.136 & 0.756$ \pm $0.013 \\ 
     Latent-Flow & 0.230$ \pm $0.029 & 0.537$ \pm $0.082 & 6.279$ \pm $0.098 & 0.760$ \pm $0.004 \\ 
     \hdashline \\ [-2.8mm]
     IVP-VAE-ODE & \textbf{0.221$ \pm $0.020} & 0.521$ \pm $0.076 & 6.383$ \pm $0.096 & 0.763$ \pm $0.048 \\
     IVP-VAE-Flow & 0.224$ \pm $0.013 & 0.519$ \pm $0.048 & \textbf{6.276$ \pm $0.104} & \underline{0.769$ \pm $0.011} \\
    \bottomrule
\end{tabular}
\endgroup
\end{table}

\subsubsection{Comparison with Imputation-Based Methods}
Another approach to work with irregular time series is to first convert them into regular time series through data imputation, and then apply conventional time series analysis methods. We also performed some experiments to compare this approach. First, we convert the used datasets into regular time series data by imputing missing values. For every missing value we impute the most recent observed value in place of the missing entry, if available. In instances where no preceding value is observed, we fill the missing position with the global mean for the respective variable within the training dataset.
We introduce three additional baseline methods \cite{GRU:Cho2014LearningPR, zhou2021informer, zeng2023transformers} used primarily for regular time series analysis and augment them with the data imputation scheme discussed above. IVP-VAE was run on the original datasets without imputation. The results of the experiments for mortality prediction in Table~\ref{tab:exp_regular_models} show that IVP-VAE outperforms missing value imputation methods. We believe this is because a large amount of data imputation will cause the original important observations to be buried in redundant information, resulting in less distinctive samples.

\begin{table}[ht]
\centering
\caption[skip=50pt]{Comparison of ODE-based methods and baselines for regular time series with missing value imputation.}
\label{tab:exp_regular_models}
\begingroup
\scriptsize
\renewcommand{\arraystretch}{0.9}

\begin{tabular}{lccc}
    \toprule
     & \multicolumn{1}{c}{MIMIC-IV} & \multicolumn{1}{c}{PhysioNet 2012} & \multicolumn{1}{c}{eICU} \\
     \cmidrule(lr){2-2} \cmidrule(lr){3-3} \cmidrule(lr){4-4}
     & \multicolumn{1}{c}{AUROC} & \multicolumn{1}{c}{AUROC} & \multicolumn{1}{c}{AUROC} \\
     \midrule
     GRU & 0.720$ \pm $0.016 & 0.709$ \pm $0.023 & 0.725$ \pm $0.012 \\ 
     Informer & 0.755$ \pm $0.015 & 0.738$ \pm $0.028 & 0.747$ \pm $0.029 \\ 
     DLinear & 0.749$ \pm $0.011 & 0.733$ \pm $0.029 & 0.741$ \pm $0.035 \\ 
     \hdashline \\ [-2.8mm]
     IVP-VAE-ODE & 0.802$ \pm $0.007 & 0.770$ \pm $0.025 & \textbf{0.786$ \pm $0.019} \\
     IVP-VAE-Flow & \textbf{0.805$ \pm $0.005} & \textbf{0.771$ \pm $0.030} & 0.786$ \pm $0.017 \\
    \bottomrule
\end{tabular}
\endgroup
\end{table}

\end{document}